\documentclass[num-refs]{wiley-article}

\usepackage{siunitx}

\papertype{Submitted to MagneticResonance in Medicine}
\paperfield{}

\title{Method for motion artifact reduction using a convolutional neural network for dynamic contrast enhanced MRI of the liver}

\author[1\authfn{1}]{Daiki Tamada, PhD}
\author[1\authfn{2}]{Marie-Luise Kromrey, MD}
\author[1\authfn{2}]{Hiroshi Onishi, MD PhD}
\author[1\authfn{2}]{Utaroh Motosugi, MD PhD}

\contrib[\authfn{1}]{Equally contributing authors.}

\affil[1]{Department of Radiology, University of Yamanashi, Chuo, Yamanashi, 409-3898, Japan}

\corraddress{Daiki Tamada PhD, Department of Radiology, University of Yamanashi, Chuo, Yamanashi, 409-3898, Japan}
\corremail{dtamada@yamanashi.ac.jp}

\presentadd[\authfn{1}]{Department of Radiology, University of Yamanashi, Chuo, Yamanashi, 409-3898, Japan}

\fundinginfo{JSPS KAKENHI, The Ministry of Education,Culture,Sports,Science and Technology of Japan, Grant/Award Number: 18K18364  \\Word count: 4095 \\No.of Figures: 8\\No. of Tables: 0}

\runningauthor{Daiki Tamada et al.}

\begin{document}

\maketitle

\begin{abstract}

\textbf{Purpose}: To improve the quality of images obtained via dynamic contrast-enhanced MRI (DCE-MRI) that include motion artifacts and blurring using a deep learning approach.\\
\textbf{Methods}: A multi-channel convolutional neural network (MARC) based method is proposed for reducing the motion artifacts and blurring caused by respiratory motion in images obtained via DCE-MRI of the liver. The training datasets for the neural network included images with and without respiration-induced motion artifacts or blurring, and the distortions were generated by simulating the phase error in k-space. Patient studies were conducted using a multi-phase T1-weighted spoiled gradient echo sequence for the liver containing breath-hold failures during data acquisition. The trained network was applied to the acquired images to analyze the filtering performance, and the intensities and contrast ratios before and after denoising were compared via Bland–Altman plots.\\
\textbf{Results}: The proposed network was found to significantly reduce the magnitude of the artifacts and blurring induced by respiratory motion, and the contrast ratios of the images after processing via the network were consistent with those of the unprocessed images.\\
\textbf{Conclusion}: A deep learning based method for removing motion artifacts in images obtained via DCE-MRI in the liver was demonstrated and validated.

\end{abstract}

\section{Introduction}

Dynamic contrast enhanced MRI (DCE-MRI) in the liver is widely used for detecting hepatic lesions and in distinguishing malignant from benign lesions. However, such images often suffer from motion artifacts due to unpredictable respiration, dyspnea, or mismatches in k-space caused by rapid injection of the contrast agent\cite{motosugi2015investigation}\cite{davenport2013comparison}. In DCE-MRI, a series of T1-weighted MR images is obtained after the intravenous injection of a gadolinium-based MR contrast agent, such as gadoxetic acid. However, acquiring appropriate data sets for DCE arterial phase MR images is difficult due to the limited scan time available in the first pass of the contrast agent. Furthermore, it has been reported that transient dyspnea can be caused by gadoxetic acid at a non-negligible frequency \cite{motosugi2015investigation}\cite{davenport2013comparison}, which results in degraded image quality due to respiratory motion-related artifacts such as blurring and ghosting\cite{stadler2007artifacts}. Especially, coherent ghosting originating from the anterior abdominal wall decrease diagnostic performance of the images\cite{chavhan2013abdominal}

 Recently, many strategies have been proposed to avoid motion artifacts in DCE-MRI. Of these, fast acquisition strategies using compressed sensing may provide the simplest way to avoid motion artifacts in the liver \cite{vasanawala2010improved}\cite{zhang2014clinical}\cite{jaimes2016strategies}. Compressed sensing is an acquisition and reconstruction technique based on the sparsity of the signal, and the k-space undersampling results in a shorter scan time. Zhang et al. demonstrated that DCE-MRI with a high acceleration factor of 7.2 using compressed sensing provides significantly better image quality than conventional parallel imaging \cite{zhang2014clinical}. Other approaches include data acquisition without breath holding (free-breathing method) using respiratory triggering and respiratory triggered DCE-MRI, which is an effective technique to reduce motion artifacts in cases of patients who are unable to suspend their respiration \cite{vasanawala2010navigated}\cite{chavhan2013abdominal}.In these approaches, sequence acquisitions are triggered based on respiratory tracings or navigator echoes, and typically provide a one-dimensional projection of abdominal images. Shreyas et al. found that the image quality in acquisitions with navigator echoes under free-breathing conditions is significantly improved. Although triggering based approaches successfully reduce the motion artifacts, it is not possible to appropriately time arterial phase image acquisition due to the long scan times required to acquire an entire dataset. In addition, miss-triggers often occur in cases of unstable patient respiration, which causes artifacts and blurring of the images. Recently, a radial trajectory acquisition method with compressed sensing was proposed  \cite{feng2014golden}\cite{feng2016xd}, which enables high-temporal resolution imaging without breath holding in DCE-MRI. However, the image quality of radial acquisition without breath holding is worse than that with breath holding even though the clinical usefulness of radial trajectory acquisition has been demonstrated in many papers \cite{chandarana2011free}\cite{chandarana2013free}.\cite{chandarana2014free}


Post-processing artifact reduction techniques using deep learning approaches have been also been proposed. Deep learning, which is used in complex non-linear processing applications, is a machine learning technique that relies on a neural network with a large number of hidden layers. Han et al. proposed a denoising algorithm using a multi-resolution convolutional network called “U-Net” to remove the streak artifacts induced in images obtained via radial acquisition \cite{han2018deep}.In addition, aliasing artifact reduction has been demonstrated in several papers as an alternative to compressed sensing reconstruction \cite{lee2017deep}\cite{yang2018dagan}\cite{hyun2018deep}. The results of several feasibility studies of motion artifact reduction in the brain \cite{Karsten2018ismrm}\cite{Patricia2018ismrm}\cite{Kamlesh2018ismrm}, abdomen \cite{Daiki2018ismrm}, and cervical spine \cite{Hongpyo2018ismrm} have also been reported. Although these post-processing techniques have been studied extensively, no study has ever demonstrated practical artifact reduction in DCE-MRI of the liver.

In this study, a motion artifact reduction method was developed based on a convolutional network (MARC) for DCE-MRI of the liver that removes motion artifacts from input MR images. Both simulations and experiments were conducted to demonstrate the validity of the proposed algorithm.

\section{Methods}

\subsection{Network architecture}

\begin{figure}[bt]
\centering
\includegraphics[width=12cm]{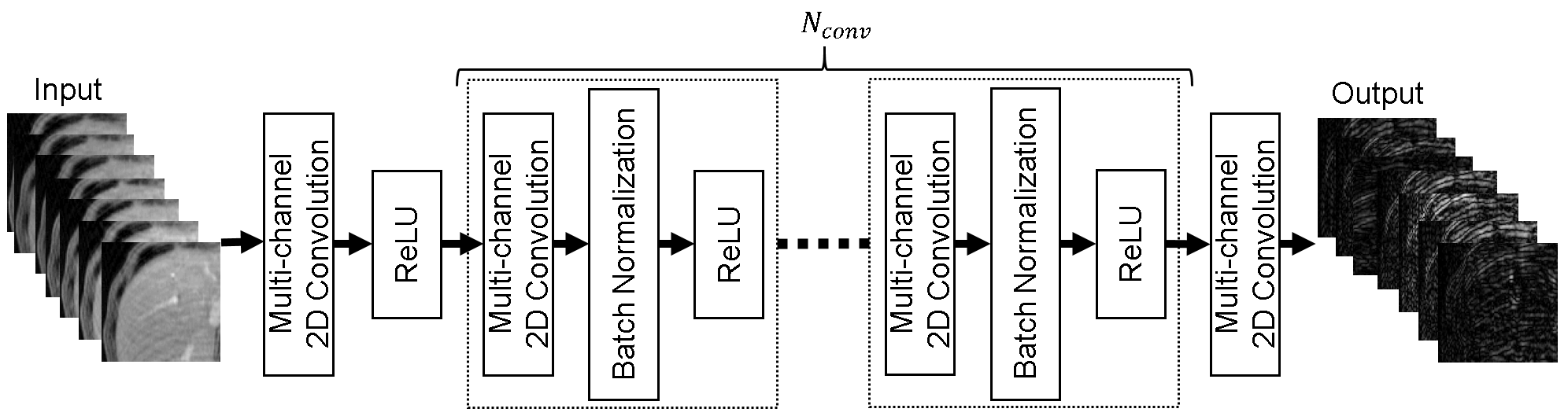}
\caption{Network architecture for the proposed convolutional neural network, two-dimensional convolutions, batch normalizations, and ReLU. The network predicts the artifact component from an input dataset. The network number of convolution layers in the network was determined by simulation-based method.}
\label{fig_network}
\end{figure}

In this paper, a patch-wise motion artifact reduction method that employs a convolutional neural network with multi-channel images (MARC) is proposed, as shown in Fig. \ref{fig_network}, which is based on the network originally proposed by Zhang et al. for Gaussian denoising, JPEG deblocking, and super-resolution of natural images\cite{zhang2017beyond}. Patch-wise training has advantages in extracting large training datasets from limited images, and efficient memory usage on host PCs and GPUs. Residual learning approach was adopted to achieve effective training of the network\cite{he2016deep}. The network relies on two-dimensional convolutions, batch normalizations, and rectified linear units (ReLU) to extract artifact components from images with artifacts. To utilize the structural similarity of the multi-contrast images, a seven-layer patched image with varying contrast, was used as input to the network. Sixty-four filters using a kernel size of 3$\times$3$\times$7 in ReLU were adopted to facilitate non-linear operation. The number of convolution layers $N_{conv}$ was determined as shown in the Analysis subsection. In the last layer, seven filters with a kernel size of 3$\times$3$\times$64 were used for second to the last layers. Finally, a 7-channel image was predicted as the output of the network. The total number of parameters was 268,423. Artifact-reduced images could then be generated by subtracting the predicted image from the input.

\subsection{Imaging}

Following Institutional Review Board approval, patient studies were conducted. This study retrospectively included 31 patients (M/F, mean age 59, range 34–79 y.o.) who underwent DCE-MRI of the liver in our institution . MR images were acquired using a 3T MR750 system (GE Healthcare, Waukesha, WI); a whole-body coil and 32-channel torso array were used for RF transmission and receiving, and self-calibrated parallel imaging (ARC) was used with an acceleration factor of 2 $\times$ 2. A three-dimensional (3D) T1-weighted spoiled gradient echo sequence with a dual-echo bipolar readout and variable density Cartesian undersampling (DISCO: differential subsampling with cartesian ordering) was used for the acquisition \cite{saranathan2012differential}, along with an elliptical-centric trajectory with pseudo-randomized sorting in ky-kz. DIXON-based reconstruction method was used to suppress fat signals\cite{reeder2004multicoil}. A total of seven temporal phase images, including pre-contrast and six arterial phases, were obtained using gadolinium contrast with end-expiration breath-holdings of 10 and 21 s. The standard dose (0.025 mmol/kg) of contrast agent (EOB Primovist, Bayer Heathcare, Osaka, Japan) was injected at the rate of 1 ml/s followed by a 20-mL saline flush using a power injector. The arterial phase scan was started 30 s after the start of the injection. The acquired k-space datasets were reconstructed using a view-sharing approach between the phases and a two-point Dixon method to separate the water and fat components. The following imaging parameters were used: flip angle = 12 $^\circ$, receiver bandwidth = $\pm$ 167 kHz, TR = 3.9 ms, TE = 1.1/2.2 ms, acquisition matrix size = 320 $\times$ 192, FOV = 340 $\times$ 340 mm$^2$, the total number of slices = 56, slice thickness = 3.6 mm. The acquired images were cropped to a matrix size of 320 $\times$ 280 after zero-filling to 320 $\times$ 320.

\subsection{Respiration-induced noise simulation}\label{sec_resp_sim}

\begin{figure}[bt]
\centering
\includegraphics[width=11cm]{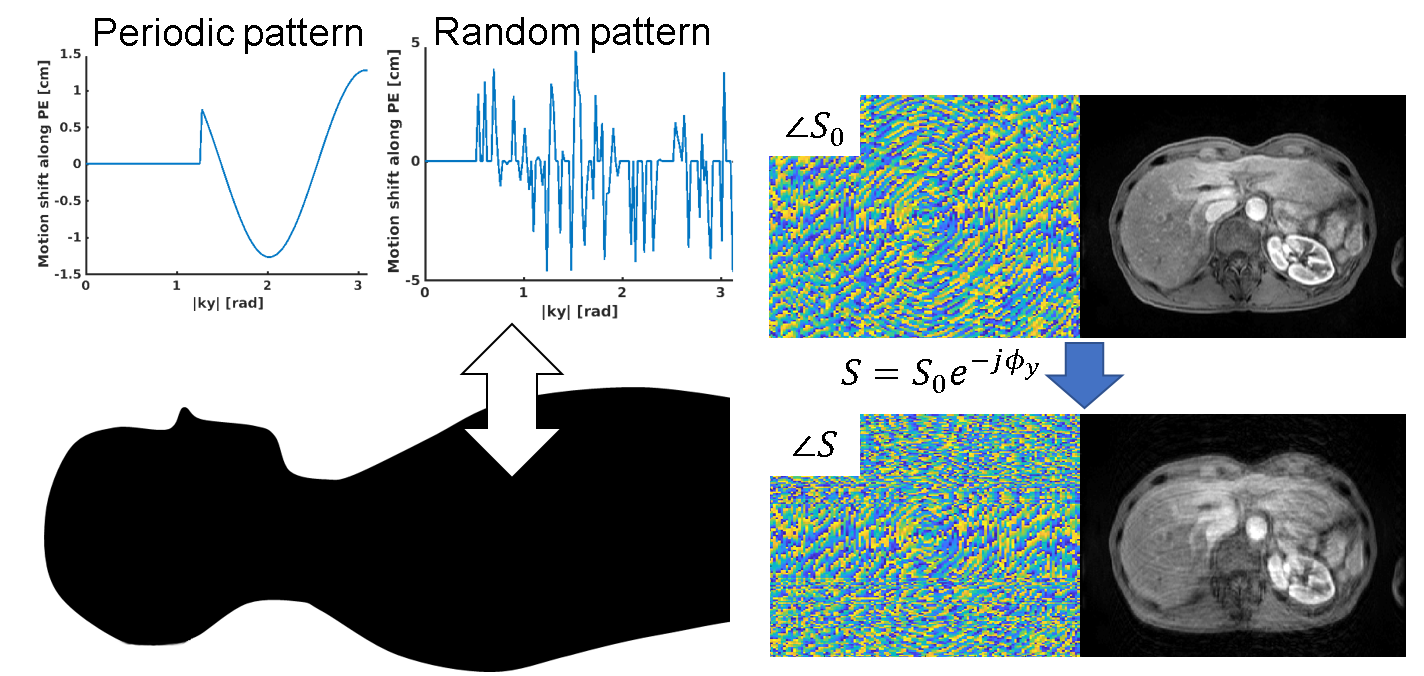}
\caption{(left) Example of a simulation of the respiratory motion artifact by adding phase errors along the phase-encoding direction in k-space. (Right) The k-space and image datasets before and after adding simulated phase errors.}
\label{fig_motion}
\end{figure}

A respiration-induced artifact was simulated by adding simulated errors to the k-space datasets generated from the magnitude-only image. Generally, a breath-holding failure causes phase errors in the k-space, which results in the artifact along the phase-encoding direction. In this study, for simplicity, rigid motion along the anterior-posterior direction was assumed, as shown in Fig. \ref{fig_motion}. In this case, the phase error was induced in the phase-encoding direction, and was proportional to the motion shift. Motion during readout can be neglected because it is performed within a millisecond order.  Then, the in-phase and out-of-phase MR signal with phase error $\phi$ can be expressed as follows:

\begin{eqnarray}
	S'_I(k_x, k_y) &=& S_I (k_x, k_y) e^{-j\phi(k_y)}\\
 	S'_O(k_x, k_y) &=& S_O (k_x, k_y) e^{-j\phi(k_y)},
\end{eqnarray}
where $S_I$ and $S_O$ are the in-phase and out-of-phase signals, respectively, without the phase error; $S'_I$ and $S'_O$ are the corresponding signals with the phase error, and $k_x$, $k_y$ represent the k-space ($-\pi < k_x < \pi$, $-\pi < k_y < \pi$) in the readout and the phase-encoding directions, respectively. Finally, k-space of the water signal ($S_W$) with the phase error can be expressed as follows:

\begin{eqnarray}
	S_W &=& \frac{S'_I+S'_O}{2}\\
 	 	&=& \frac{S_I+S_O}{2}e^{-j\phi(k_y)}\\
    	&=& \mathcal{F}[I_W]e^{-j\phi(k_y)},
\end{eqnarray}
where $\mathcal{F}$ is the Fourier operator, and Iw denotes the water image. It is clear from the above equation that artifact simulation can be implemented by simply adding the phase error components to the k-space of the water image. In this study, the k-space datasets were generated from magnitude-only water images. To simulate the background $B_0$ inhomogeneity, the magnitude images were multiplied by $B_0$ distributions derived from polynomial functions up to the third order. The coefficients for the functions were determined randomly so that the peak-to-peak value of the distribution was within $\pm5$ ppm ($\pm4.4$ radian)

To generate a motion artifact in the MR images, we used two kinds of phase error patterns: periodic and random. Generally, severe coherent ghosting artifacts are observed along the phase-encoding direction. Although there are several factors that generate artifacts in the acquired images during DCE-MRI including respiratory, voluntary motion, pulsatile arterial flow, view-sharing failure, and unfolding failure\cite{stadler2007artifacts}\cite{arena1995mr}, the artifact from the abdominal wall in the phase-encoding direction is mainly recognizable. In the case of centric-order acquisitions, the phase mismatching in the k-space results in high-frequency and coherent ghosting. An error pattern using simple sine wave with random frequency, phase, and duration was used to simulate the ghosting artifact. It was assumed that motion oscillations caused by breath-hold failures occurred after a delay as the scan time proceeded. The phase error can be expressed as follows:

\[
  \phi(k_y) = \left\{ \begin{array}{ll}
    0 & (|k_y| < k_{y0} ) \\
    2\pi \frac{k_y \Delta sin(\alpha k_y + \beta)}{N} & (otherwise),
  \end{array} \right.
\]
where $\Delta$ denotes the significance of motion, $\alpha$ is the period of the sine wave which determines the frequency, $\beta$ is the phase of the sine wave, and $k_{y0}, (0 < k_{y0} < \pi)$ is the delay time for the phase error. In this study, the values of $\Delta$ (from 0 to 20 pixels, which equals 2.4-2.6 cm depending on FOV), $\alpha$ (from 0.1 to 5 Hz), $\beta$ (from 0 to $\pi /4$) and $k_{y0}$ (from $\pi/10$ to $\pi/2$) were selected randomly. The period $\alpha$ was determined such that it covered the normal respiratory frequency for adults and elderly adults, which is generally within 0.2-0.7 Hz\cite{rodriguez2013normal}. In addition to the periodic noise, random phase error pattern was also used to simulate non-periodic irregular motion as follow. First, the number of phase-encoding lines, which have the phase error, was randomly determined as between 10-50 \% of all phase-encoding lines except at the center region of the k-space ($-\pi /10 < k_{y0} < \pi /10$). Then, the significance of the error was determined randomly line-by-line in the same manner as used for periodic noise.

\subsection{Network Training}

The processing was implemented in MATLAB 2018b on a workstation running Ubuntu 16.04 LTS with an Intel Xeon CPU E5-2630, 128 GB DDR3 RAM, and an NVIDIA Quadro P5000 graphics card.

\begin{figure}[bt]
\centering
\includegraphics[width=12cm]{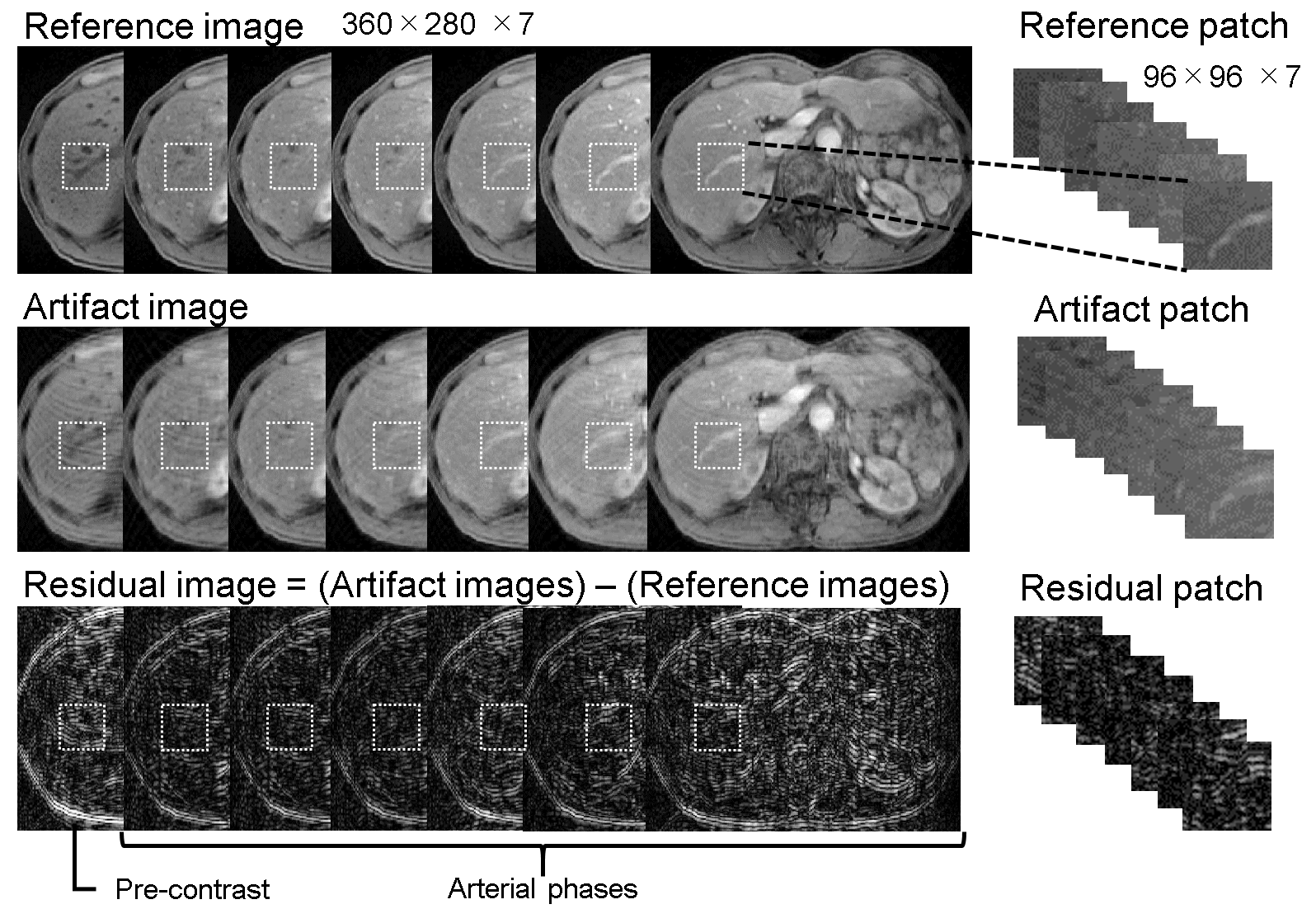}
\caption{Data processing for the training. The artifact images were simulated from the reference images. Residual images were calculated by subtracting the reference patches from the artifact patches. A total of 125,273 patches were generated by randomly cropping small images from the original images.}
\label{fig_dataset}
\end{figure}

The data processing sequence used in this study is summarized in Fig \ref{fig_dataset}. Training datasets containing artifacts and residual patches were generated using multi-phase magnitude-only reference images (RO $\times$ PE $\times$ SL $\times$ Phase: 320 $\times$ 280 $\times$ 56 $\times$ 7) acquired from six patients selected by a radiologist from among the 26 patients in the study. The radiologist confirmed that all reference images were successfully acquired without motion artifacts. For the multi-phase slices (320 $\times$ 280 $\times$ 7) of the images, 125,273 patches 48 $\times$ 48 $\times$ 7 in size were randomly cropped. The resulting patches that contained only background signals were removed from the training datasets. Images with motion artifact (artifact images) were generated using the reference images, as explained in the previous subsection. Artifact patches, which were used as inputs to the MARC, were cropped from the artifact images using the same method as that for the reference patches. Finally, residual patches, which were used for the output of the network, were generated by subtracting the reference patches from the artifact patches. All patches were normalized by dividing them by the maximum value of the artifact images.

Network training was performed using a Keras and Tensorflow backend (Google, Mountain View, CA), and the network was optimized using the Adam algorithm with a learning rate of 0.001. The optimization was conducted with a mini-batch of 64 patches. A total of 100 epochs with an early-stopping patience of 10 epochs were completed for convergence purposes and the L1 loss function was used as the residual components between the artifact patches and outputs were assumed to be sparse.

\begin{equation}
Loss(I_{art}, I_{out}) = \frac{1}{N}\sum_i^N \| I_{art} - I_{out} \|_1,
\end{equation}
where $I_{art}$ represents the artifact patches, $I_{out}$ represents the outputs predicted using the MARC, and N is the number of data points. Validation for L1 loss was performed using K-fold cross validation (K = 5).

The $N_{conv}$ used in the network was determined by maximizing the structural similarity (SSIM) index between the reference and artifact-reduced patches of the validation datasets. Here, the SSIM index is a quality metric used for measuring the similarity between two images, and is defined as follows:

\begin{equation}
SSIM(I_{ref}, I_{den}) = \frac{(2\mu_{ref} \mu_{den}+c_1)(2\sigma_{ref,den}+c_2)}{(\mu_{ref}^2 + \mu_{den}^2 + c_1)(\sigma_{ref}^2 + \sigma_{den}^2+c_2)},
\end{equation}
where, $I_{ref}$ and $I_{den}$ are input and artifact-reduced patches, $\mu$ is the mean intensity, $\sigma$ denotes the standard deviation, and $c_1$ and $c_2$ are constants. In this study, the values of $c_1$ and $c_2$ were as described in \cite{wang2004image}.

The number of patients used for the training versus the L1 loss with 100-epoch training was plotted to investigate the relationship between the size of the training datasets and training performance. The average sample size for one patient was 7916 training patches and 3417 validation patches for 11,333 patches. A validation dataset of 37,582 patches used for these trainings was generated from the 11 patients.

\subsection{Analysis}\label{sec_anal}

To demonstrate the performance of the MARC to reduce the artifacts in the DCE-MR images acquired during unsuccessful breath holding, the following experiments were conducted using the data from the 20 remaining patients in the study. To identify biases in the intensities and liver-to-aorta contrast between the reference and artifact-reduced images, a Bland–Altman analysis, which plots the differences of the two images versus their average, was used in which the intensities were obtained from the central slice in each phase. The Bland–Altman analysis for the intensities was conducted in the subgroups of high (mean intensity $\geq$ 0.46) and low (mean intensity < 0.46) intensities. For convenience, half of the maximum mean intensity (0.46) was used as the threshold. The mean signal intensities of the liver and aorta were measured by manually placing the region-of-interest (ROI) on the MR images, and the ROI of the liver was carefully located in the right lobe to exclude vessels. The same ROIs were applied to all other phases of the images. The quality of images before and after applying the MARC were visually evaluated by a radiologist (M.K.) with three years of experience in abdominal radiology who was not told whether each image came before or after the MARC was applied. The radiologist evaluated the images using a 5-point scale based on the significance of the artifacts (1 = no artifact; 2 = mild artifacts; 3 = moderate artifacts; 4 = severe artifacts; 5 = non-diagnostic). The scores of more than 1 were analyzed statistically by using the Wilcoxon signed rank test. To confirm the validity of the anatomical structure after applying the MARC, the artifact-reduced images in the arterial phase were compared with those without the motion artifact, which were obtained from separate MR examinations performed 71 days apart in the same patients. The same sequence and imaging parameters were used for the acquisition.


\section{Results}

\begin{figure}[bt]
\centering
\includegraphics[width=7cm]{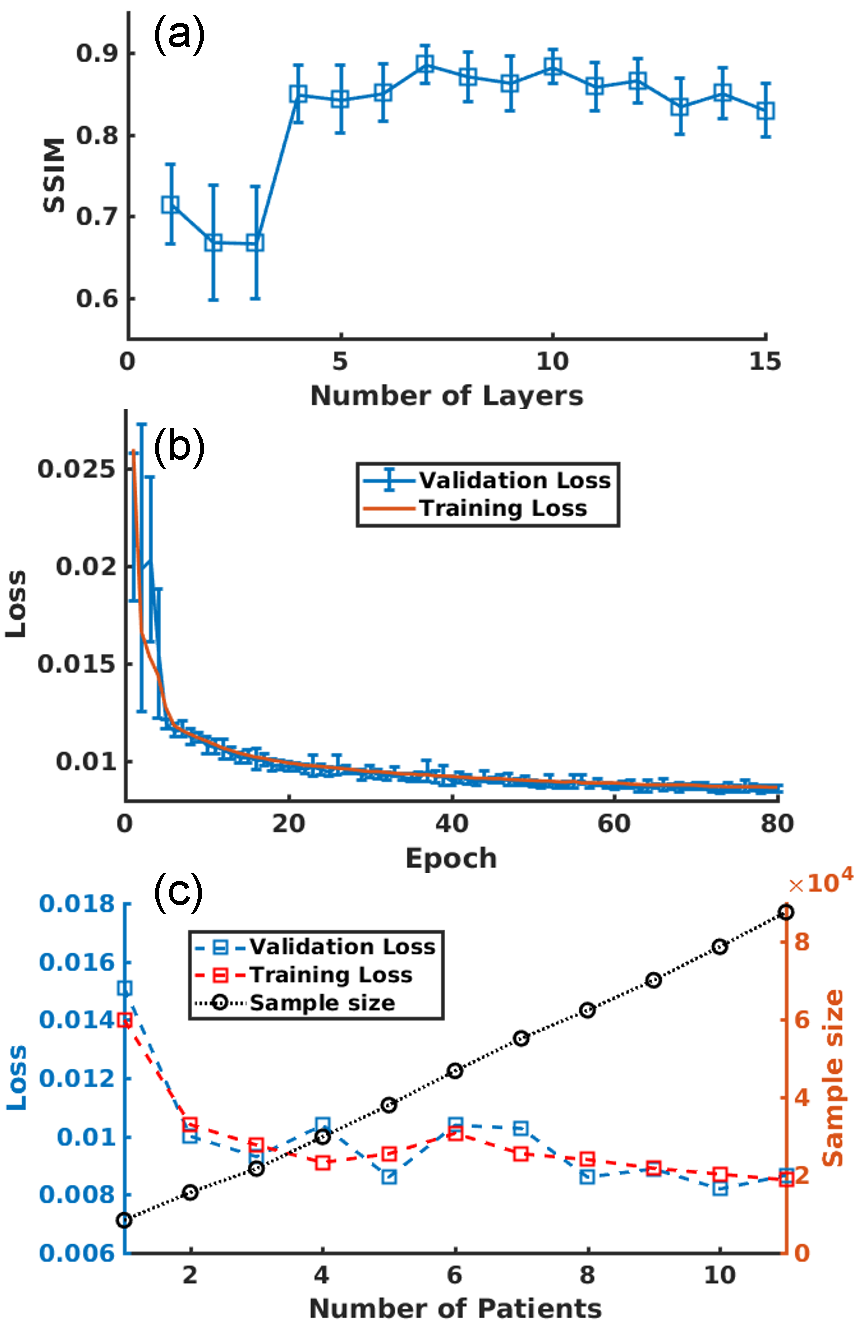}
\caption{(a) SSIM changes depending on the number of layers ($N_{conv}$). The highest SSIM (0.89) was obtained with an $N_{conv}$ of 7. (b) The L1 loss decreased in both the training, and validation datasets as the number of epochs increased. No further decrease was visually observed in > 70 epochs. The training was terminated by early stopping in 80 epochs.Error bars on the validation loss represent the standard deviation for K-fold cross validation. (c) Validation loss, training loss, and sample size were plotted against the number of patients. Smaller loss was observed as the sample size and number of patients increased.}
\label{fig_SSIM}
\end{figure}

\begin{figure}[bt]
\centering
\includegraphics[width=11cm]{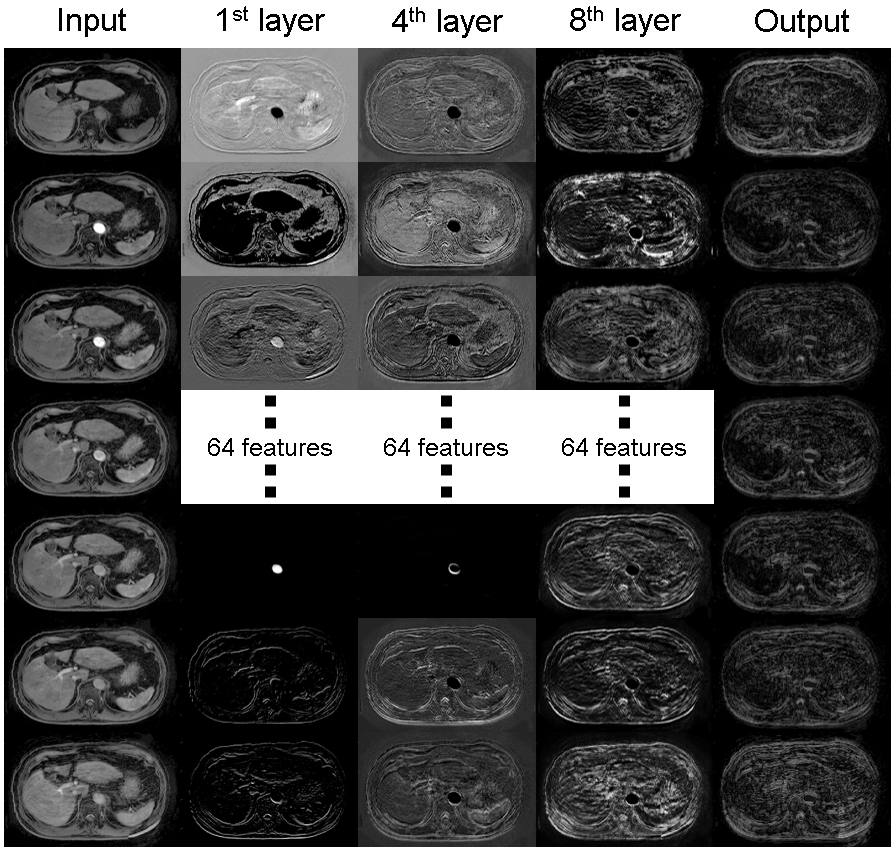}
\caption{Features extracted from 1st, 4th, and 8th layers of the developed network corresponding to specific input and output. Low- and high-frequency components were observed in the lower layers. On the other hands, an artifact-like pattern was extracted from the higher layer.}
\label{fig_Features}
\end{figure}

The changes in the mean and standard deviation ($\mu$) of the SSIM index between the reference and artifact-reduced images are plotted against $N_{conv}$ in Fig. \ref{fig_SSIM} (a), and the results show that the network with an $N_{conv}$ of more than four exhibited a better SSIM index, while networks with an $N_{conv}$ of four or below had a poor SSIM index. In this study, an $N_{conv}$ of seven was adopted in the experiments as this value maximized the SSIM index (mean: 0.87, $\mu$: 0.05). The training was successfully terminated by early stopping in 70 epochs as shown in Fig. \ref{fig_SSIM} (b). Figure \ref{fig_SSIM} (c) shows the number of patients used for the training versus the training and validation losses, and the sample size. The results implied that stable convergence was achieved when the sample size was 3 or more although the training with few patients gave inappropriate convergence. The features using the trained network extracted from the 1st, 4th, and 8th intermediate layers corresponding to specific input and output are shown in Fig. \ref{fig_Features}. Higher frequency ghosting-like patterns were extracted from the input in the 8th layer. 

\begin{figure}[bt]
\centering
\includegraphics[width=11cm]{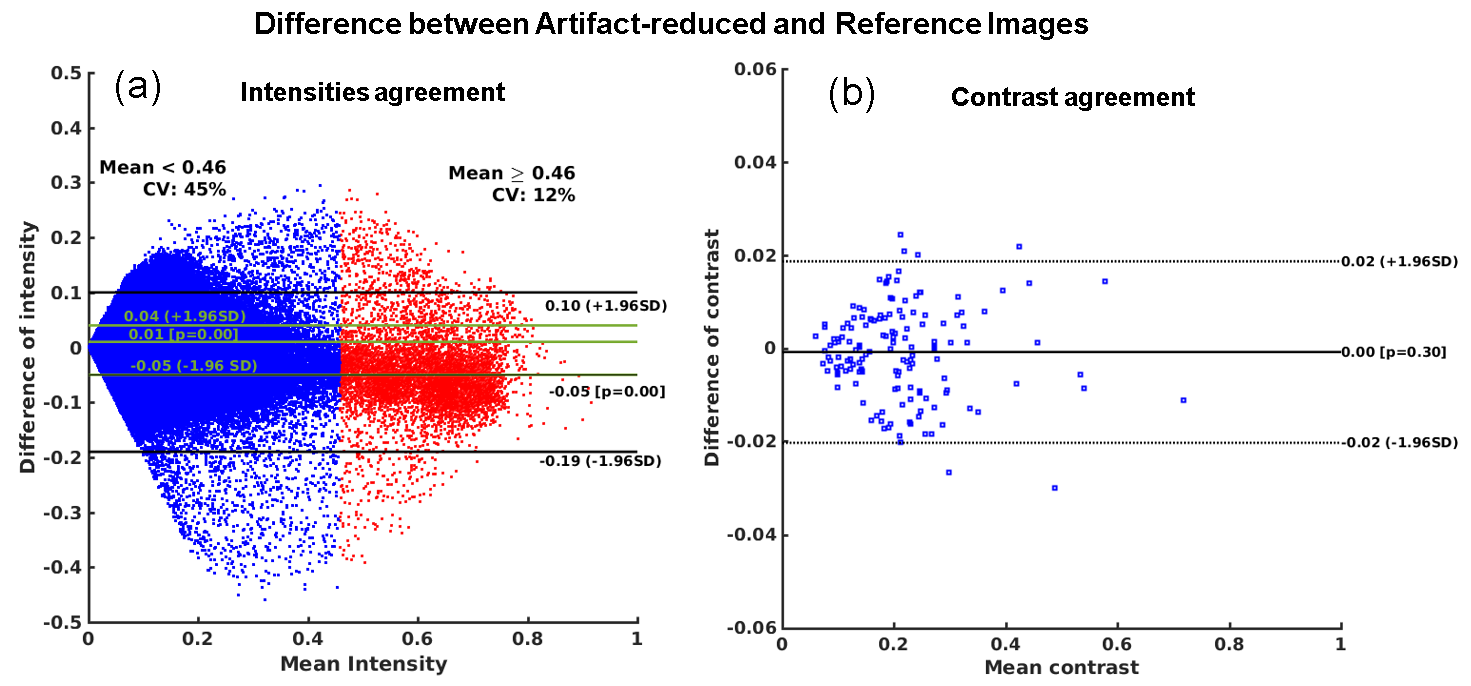}
\caption{Bland–Altman plots for (a) the intensities and (b)the liver-to-aorta contrast ratio between the reference and artifact-reduced images in the validation dataset. The mean difference in the intensities was 0.01 (95 \% CI, -0.05-0.04) in the areas corresponding to mean intensity of < 0.46 and -0.05 (95 \% CI, -0.19-0.01) in the parts of mean intensity of $\geq$ 0.4. Mead difference in the contrast ratio was 0.00 (95 \% CI, -0.02-0.02). These results indicated that there were no systematic errors in the contrast ratios, whereas the intensities of the artifact-reduced images were lower than that of the reference images due to the effect of artifact reduction especially in the area with high signal intensities. }
\label{fig_BAplot}
\end{figure}

Figures \ref{fig_BAplot} (a) and (b) show the Bland–Altman plots of the intensities and liver-to-aorta contrast ratios between the reference and artifact-reduced images. The differences in the intensities between the two images (mean difference = 0.01 (95 \% CI, -0.05-0.04) for mean intensity < 0.46 and mean difference = -0.05 (95 \% CI, -0.19-0.01) for mean intensity $\geq$ 0.46) were heterogeneously distributed, depending on the mean intensity. The intensities of the artifact-reduced images were lower than that of the references by ~15 \% on average, which can be seen in the high signal intensity areas shown in Fig. \ref{fig_BAplot} (a). A Bland–Altman plot of the liver-to-aorta contrast ratio (Fig. \ref{fig_BAplot} (b)) showed no systematic errors in contrast between the two images. 

\begin{figure}[bt]
\centering
\includegraphics[width=12cm]{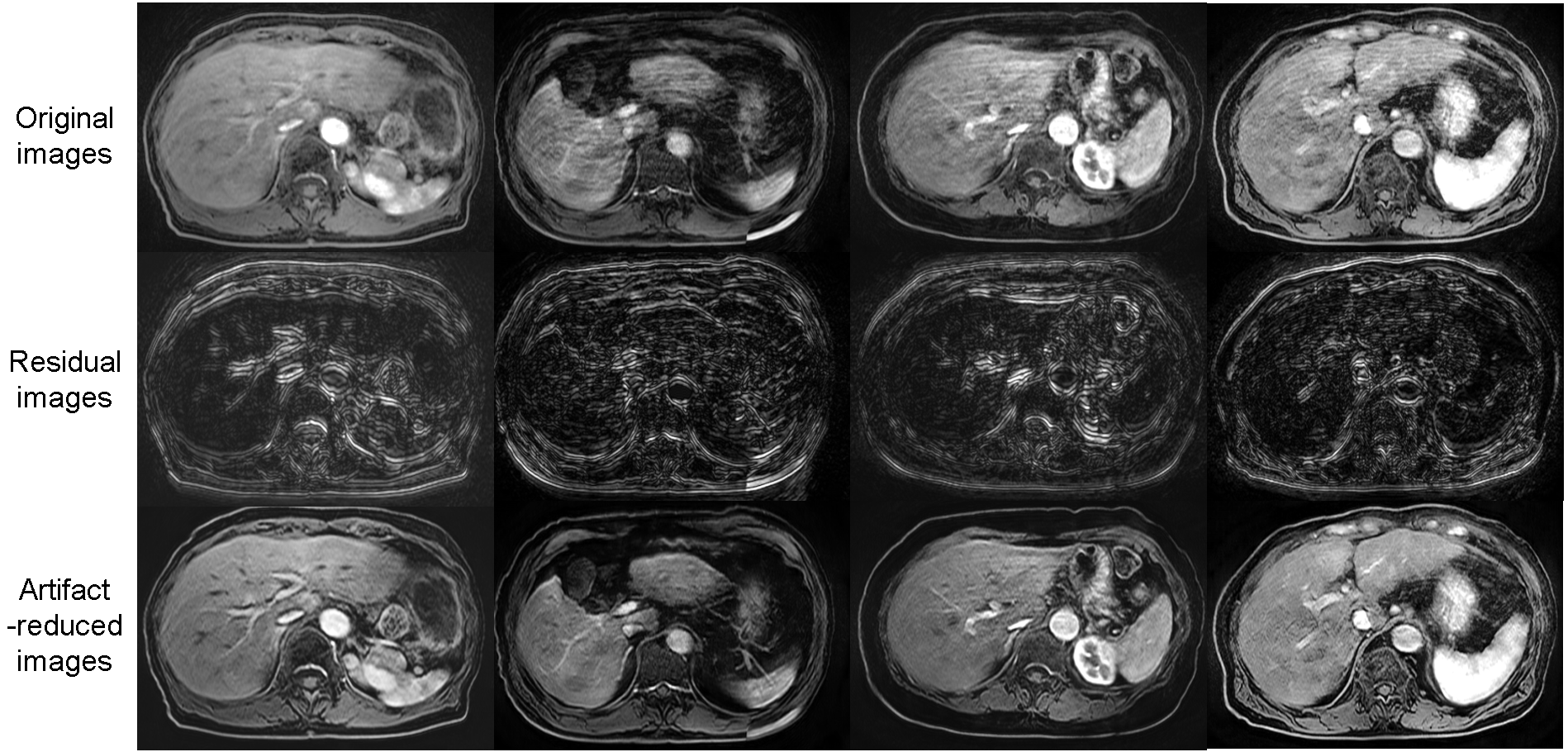}
\caption{Examples of artifact reduction with MARC for a patient from the validation dataset. The motion artifacts in the images (upper row) were reduced (lower row) by using the MARC.The residual components are shown in the middle row.}
\label{fig_results}
\end{figure}

The image quality of the artifact-reduced images (mean (SD) score = 3.2(0.63)) were significantly better (P < 0.05) than that of the reference images (mean (SD) score =2.7(0.77)), and the respiratory motion-related artifacts (Fig. \ref{fig_results} top row) were reduced by applying MARC (Fig. \ref{fig_results} bottom row). The middle row in Fig. \ref{fig_results} shows the extracted residual components for the input images.

\begin{figure}[bt]
\centering
\includegraphics[width=12cm]{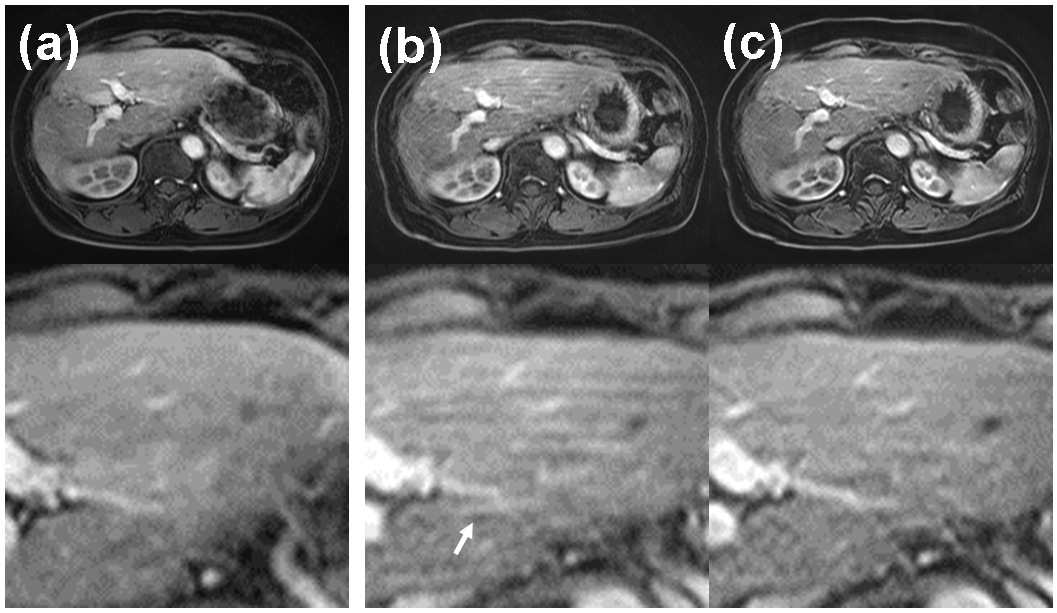}
\caption{(a, b) MR image in the arterial phase with and without breath-hold failure. (c) The artifact-reduced image of (b). The images were acquired in difference studies with same imaging parameters.}
\label{fig_Comparison}
\end{figure}

The images with and without breath-hold failure are shown in Fig. 8 (a, b). The motion artifact in Fig. \ref{fig_Comparison} (b) was partially reduced by using MARC, as shown in Fig. \ref{fig_Comparison} (c). This result indicated that there was no loss of critical anatomical details, and additional blurring was observed although moderate artifact on the right lobe remained.

\section{Discussion}

In this paper, an algorithm to reduce the number of motion-related artifacts after data acquisition was developed using a deep convolutional network, and was then used to extract artifacts from local multi-channel patch images. The network was trained using reference MR images acquired with appropriate breath-holding, and noisy images were generated by adding phase error to the reference images. The number of convolution layers in the network was semi-optimized in the simulation. Once trained, the network was applied to MR images of patients who failed to hold their breath during the data acquisition. The results of the experimental studies demonstrate that the MARC successfully extracted the residual components of the images and reduced the amount of motion artifacts and blurring. No study has ever attempted to demonstrate blind artifact reduction in abdominal imaging, although many motion correction algorithms with navigator echoes or respiratory signal have been proposed \cite{vasanawala2010navigated}\cite{brau2006generalized}\cite{cheng2012nonrigid}. In these approaches, additional RF pulses and/or longer scan time will be required to fill the k-space signal whereas MARC enables motion reduction without sequence modification or additional scan time. The processing time for one slice was 4 ms, resulting in about 650 ms for all slices of one patient. This computational cost is acceptable for practical clinical use.

In MRI of the liver, DCE-MRI is mandatory in order to identify hypervascular lesions, including hepatocellular carcinoma \cite{tang2017evidence}\cite{chen2016added}, and to distinguish malignant from benign lesions. At present, almost all DCE-MR images of the liver are acquired with a 3D gradient echo sequence due to its high spatial resolution and fast acquisition time within a single breath hold. Despite recent advances in imaging techniques that improve the image quality \cite{yang2016sparse}\cite{ogasawara2017image}, it remains difficult to acquire uniformly high quality DCE-MRI images without respiratory motion-related artifacts. In terms of reducing motion artifacts,  the unpredictability of patients is the biggest challenge to overcome, as the patients who will fail to hold their breath are not known in advance. One advantage of the proposed MARC algorithm is that it is able to reduce the magnitude of artifacts in images that have been already acquired, which will have a significant impact on the efficacy of clinical MR.

In the current study, an optimal $N_{conv}$ of seven was selected based on the SSIM indexes of the reference image and the artifact-reduced image after applying MARC. The low SSIM index observed for small values of $N_{conv}$was thought to be due to the difficulty of modeling the features of the input datasets with only a small number of layers. On the other hand, a slight decrease in the SSIM index were observed for $N_{conv}$ of >12. This result implies that overfitting of the network occurred by using too many layers. To overcome this problem, a larger number of learning datasets and/or regularization and optimization of a more complicated network will be required.

Several other network architectures have been proposed for the denoising of MRI images. For example, U-Net \cite{ronneberger2015u}, which consists of upsampling and downsampling layers with skipped connections, is a widely used fully convolutional network for the segmentation \cite{dalmics2017using}, reconstruction, and denoising\cite{yu2017deep} of medical images. This architecture, which was originally designed for biomedical image segmentation, uses multi-resolution features instead of a max-pooling approach to implement segmentation with high localization accuracy. Most of the artifacts observed in MR images, such as motion, aliasing, or streak artifacts, are distributed globally in the image domain because the noise and errors usually contaminate the k-space domain. It is known that because U-Net has a large receptive field, these artifacts can be effectively removed using global structural information. Generative adversarial networks (GANs) \cite{goodfellow2014generative}, which are comprised of two networks, called the generator and discriminator, is another promising approach for denoising MR images. Yang et al. proposed a network to remove aliasing artifacts in compressed sensing MRI using a GAN-based network with a U-Net generator\cite{yang2018dagan}. We used patched images instead of a full-size image, because it was difficult to implement appropriate training with limited number of datasets as well as owing to computational limitation. However, we believe this approach is reasonable because the pattern of artifact due to respiratory motion looks similar in every patch, even though the respiratory artifact is distributed globally. Although it should be studied further in the future, we consider that MARC from the patched image can be generalized to a full-size image from our results. Recently, the AUtomated TransfOrm by Manifold APproximation (AUTOMAP) method, which uses full connection and convolution layers, has been proposed for MRI reconstruction \cite{zhu2018image}. The AUTOMAP method directly transforms the domain from the k-space to the image space, and thus enables highly flexible reconstruction for arbitrary k-space trajectories. Three-dimensional CNNs , which are network architectures for 3D images \cite{kamnitsas2017efficient}\cite{chen2018efficient} are also promising method. However, these networks require large number of parameters, huge memory on GPUs and host computers, and long computational time for training and hyperparameter tuning. Therefore, it is still challenging to apply these approaches in practical applications. These network architectures may be combined to achieve more spatial and temporal resolution. It is anticipated that further studies will be conducted on the use of deep learning strategies in MRI. 

The limitations in the current study were as follows. First, clinical significance was not fully assessed. While the image quality appeared to improve in almost all cases, it will be necessary to confirm that no anatomical/pathological details were removed by MARC before this approach can be clinically applied. Second, simple centric acquisition ordering was assumed when generating the training datasets, which means that MARC can only be applied for a limited sequence. Additional training will be necessary before MARC can be generalized to more pulse sequences. In addition, realistic simulation can offer further improvement of our algorithm because noise simulation in this study was based on the assumption that ghosting originates from simple rigid motion. Moreover, the artifact was simulated in the generated k-space data from images for clinical use. Simulation in the original k-space data may offer different results. We need further researches to reveal which approach would be appropriate for artifact simulation.

The research on diagnostic performance using deep learning-based filters has not been performed sufficiently in spite of considerable effort spent for the development of algorithms. Our approach can provide additional structures and texture to the input images using the information learned from the trained datasets. Therefore, no essential information was added using MARC although image quality based on visual assessment was improved. However, even non-essential improvement may help non-expert or inexperienced readers to find lesions in the images. Further research on diagnostic performance will be required to demonstrate its clinical usefulness.

\section{Conclusion}

In this study, a deep learning-based network was developed to remove motion artifacts in DCE-MRI images. The results of experiments showed the proposed network effectively removed the motion artifacts from the images. These results indicate the deep learning-based network has the potential to also remove unpredictable motion artifacts from images.

\bibliography{refs}

\end{document}